%% file: main.tex
\documentclass[]{style}

\usepackage[toc,page,header]{appendix}
\usepackage{enumitem}
\usepackage{xcolor}
\usepackage[dvipsnames]{xcolor}
\usepackage{listings}
\usepackage{caption}

\input{resources/packages}

\AtBeginDocument{%
}
\newcommand{\appref}[1]{\hyperref[#1]{Appendix~\ref*{#1}}}

\usepackage{needspace}
\let\uwdsection\section
\renewcommand{\section}{\Needspace*{5\baselineskip}\uwdsection}

\newcommand{\boldtitle}[1]{{\bfseries #1}}
\makeatletter
\renewcommand{\title}[1]{\newcommand{\titlelist}{{\titlefont\sffamily #1\par}}}
\makeatother
\title{\boldtitle{UniWorld-Design}: From Pixel Generation\\ to Layer-Native Design}

\author{UniWorld Team\texorpdfstring{$^{1,2\ast}$}{}}

\newcommand\blfootnote[1]{%
  \begingroup
  \renewcommand\thefootnote{}\footnote{#1}%
  \addtocounter{footnote}{-1}%
  \endgroup}

\affiliation[1]{Peking University}
\affiliation[2]{Rabbitpre AI}

\newcommand{\tsystem}{UniWorld-Design}
\newcommand{\tshort}{UniWorld}
\newcommand{\ttrgba}{T2RGBA}
\newcommand{\ttil}{I2L}

\input{sec/0_abstract}

\checkdata[
\raisebox{-0.3em}{\includegraphics[width=0.025\linewidth]{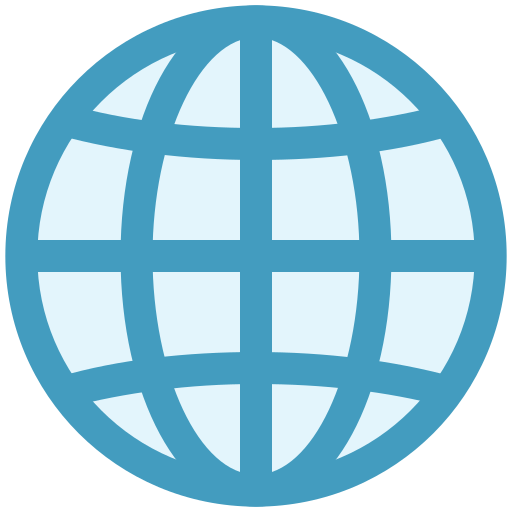}}~~Project Website]{\href{https://rabbitvis.rabbitpre.com/blog}{\texttt{https://rabbitvis.rabbitpre.com/blog}}
\\[-2.3ex]}

\begin{document}
\maketitle
\blfootnote{$^{\ast}$Full author list in
\hyperref[sec:contributions]{Contributions}.}

\vspace{-9mm}
\noindent\begin{minipage}{\linewidth}
\centering
\includegraphics[width=\linewidth]{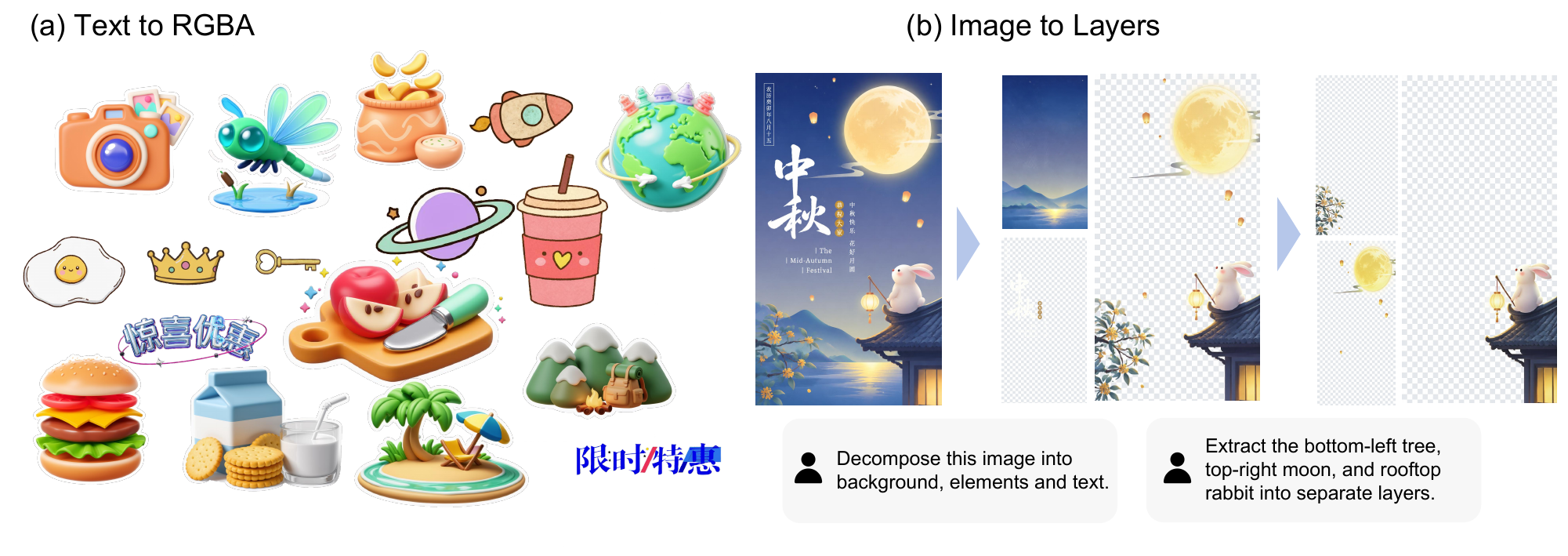}
\captionof{figure}{\small\textbf{\tsystem{} generates and decomposes design assets
natively in RGBA layers.} \textbf{(a)} \ttrgba{}: one prompt, one standalone
transparent asset. \textbf{(b)} \ttil{} follows structured prompts for
instruction-controlled semantic decomposition; any returned layer can be supplied to a
subsequent call for recursive decomposition.}
\label{fig:teaser}
\end{minipage}

\input{sec/1_intro}
\input{sec/2_related_work}
\input{sec/3_method}
\input{sec/4_experiment}
\input{sec/discussion}
\input{sec/5_conclusion}

\clearpage

\bibliographystyle{plainnat}
\setlength{\bibhang}{0pt}
\setlength\bibindent{0pt}
\bibliography{main}

\clearpage
\appendix
\input{sec/contributions}
\input{sec/6_appendix}

\end{document}

%% file: resources/packages.tex
\usepackage{natbib}

\usepackage{CJKutf8}

\usepackage{xargs}  

\usepackage{todonotes}  

\usepackage{multirow}

\usepackage{cleveref}

\usepackage{amsmath}
\usepackage{dsfont}

\usepackage{svg}

\usepackage{mathrsfs}
\usepackage{adjustbox}
\usepackage{multirow}
\usepackage{multicol}
\usepackage{tcolorbox}
\usepackage{changepage}
\usepackage{enumitem}
\usepackage{graphicx}
\usepackage{amssymb}
\usepackage{xcolor}
\usepackage{float}
\usepackage{multirow}
\usepackage{threeparttable}
\usepackage{graphicx}
\usepackage{subcaption}
\usepackage{algorithm}
\usepackage{algpseudocode}
\usepackage{wrapfig}
\usepackage[table]{xcolor}  
\usepackage{colortbl}       
\usepackage{tabularx} 
\usepackage{makecell} 
\usepackage[dvipsnames]{xcolor}

\newcolumntype{g}{>{\columncolor{gray!10}}c} 
\newcolumntype{A}{>{\raggedright\arraybackslash}p{4.5cm}} 
\newcolumntype{Y}{>{\centering\arraybackslash}X} 

\definecolor{catgray}{gray}{0.9}
\definecolor{skyblue}{rgb}{0.53,0.81,0.92} 

\colorlet{skyblue!30}{skyblue!30!white} 

\definecolor{customblue}{RGB}{70,130,180}  

\newtcolorbox{evolbox}[2][]{%
  enhanced,
  colframe=customblue,
  colback=white,
  coltitle=white,
  rounded corners,
  boxrule=1pt,
  titlerule=0pt,
  toptitle=1mm,
  bottomtitle=1mm,
  fonttitle=\bfseries,
  width=#2\textwidth, 
  #1
}

\usepackage{minitoc}
\usepackage{url}

\PassOptionsToPackage{table,xcdraw}{xcolor}
\usepackage{titletoc}
\usepackage{placeins}
\usepackage{pifont}

\definecolor{RowBlue}{HTML}{E9F2FB}
\definecolor{RowRed}{HTML}{F9EAEA}
\definecolor{Top1}{HTML}{50DB4B} 
\definecolor{Top2}{HTML}{A5FFA2} 
\definecolor{Top3}{HTML}{D9FFD9} 
\definecolor{Sub1}{HTML}{EAB8B8}
\definecolor{Sub2}{HTML}{E4E4E4}

\definecolor{gearred}{HTML}{D85140}
\definecolor{reprablue}{HTML}{5384ED}
\definecolor{steorange}{HTML}{EF8444}
\definecolor{softgreen}{HTML}{658E40}

\renewcommand{\emph}[1]{\textit{#1}}

\definecolor{codepink}{RGB}{220,20,120}
\definecolor{codegreen}{RGB}{0,150,0}
\definecolor{codegray}{RGB}{140,140,140}
\definecolor{codeorange}{RGB}{230,120,60}

\lstdefinestyle{pytorchstyle}{
    language=Python,
    basicstyle=\ttfamily\small,
    keywordstyle=\color{codepink}\bfseries,
    commentstyle=\color{codegray}\itshape,
    stringstyle=\color{codeorange},
    numberstyle=\tiny\color{codegray},
    numbers=none,
    showstringspaces=false,
    breaklines=true,
    frame=none,
    columns=fullflexible,
    keepspaces=true,
    xleftmargin=1.5em
}

%% file: sec/0_abstract.tex
\abstract{
We introduce \textbf{\tsystem{}}, a framework that redefines image generation from flat
pixel synthesis to structured visual composition, with semantic RGBA layers as the atomic
units of generation, understanding, and editing. Our key insight is that \textit{pixels define how an
image is rendered, whereas layers define how an image is created, understood, and edited}.
Just as human designers create and manipulate visual content through layers rather than raw
pixels, \tsystem{} equips multimodal generative models with a layer-native design space.
\tsystem{} comprises two models. The Text-to-RGBA (\ttrgba{}) model generates
standalone RGBA assets directly from text. The Image-to-Layer (\ttil{}) model
conditions on a finished image, a global instruction and per-layer prompts, and jointly
produces ordered, complete semantic RGBA layers. Its instruction interface supports top-level
decomposition, recursive decomposition and targeted extraction, making layering an
instruction-addressable operation for agentic editing. Because \ttil{} learns complete
semantic objects rather than visible-pixel partitions, its layers stay usable when moved or
removed. On the Crello benchmark, \ttil{} reduces per-layer RGB L1 error by $37\%$ and achieves a
$34\%$ relative improvement in Alpha Soft IoU over Qwen-Image-Layered. Separately, \ttrgba{} achieves the highest
CLIP Score, outperforming LayerDiffuse and OmniAlpha.
}

%% file: sec/1_intro.tex
\section{Introduction}
\label{sec:intro}

Text-to-image models have made remarkable progress in producing visually compelling images
from language \citep{esser2024sd3,lin2025uniworld,diao2026sensenova,Yan_2026_CVPR}, but they still largely
synthesize a flat RGB canvas.
Pixels determine how an image is rendered; they do not explicitly describe how it is
composed, which semantic elements it contains, or how those elements should be manipulated.
Human designers instead build posters, illustrations and social media graphics as ordered
layers---backgrounds, subjects, decorations and text---that can be independently selected,
moved, recolored, replaced or removed
\citep{yamaguchi2021canvasvae,suzuki2025layerd}. The limitation of a flat output is therefore
not only that editing is difficult, but that editable structure is absent from the output
space itself.

Once a layered design is flattened into a raster image, its editable structure is lost:
individual objects, their layer order and hierarchy, and content hidden by occlusion are no
longer stored explicitly. Any subsequent attempt to edit a person, a text block or the
background must first reconstruct that structure from pixels, and this recovery is generally
incomplete. Whole-image instruction editing may alter content that should remain unchanged
\citep{brooks2023instructpix2pix,yin2025qwenimagelayered}; masks can localize edits
\citep{couairon2023diffedit}; and foreground segmentation can extract visible regions
\citep{zheng2024birefnet,qin2022dis}. However, these methods do not reconstruct the original
layer stack or content hidden by occlusion
\citep{zhu2017semanticamodal,ozguroglu2024pix2gestalt}. The limitation therefore lies not only
in segmentation accuracy, but also in the absence of editable structure from the raster
representation itself. PSD-like workflows provide the natural reference: each element is
stored from the outset as an independently editable object with an explicit place in the
composition.

Recent image-to-layer models, including Qwen-Image-Layered and OmniPSD, demonstrate that
editable layer structures can be recovered from finished images
\citep{yin2025qwenimagelayered,liu2025omnipsd}. Agentic editors likewise show the value of
planning and structured execution \citep{wang2024genartist,ye2026agentbanana,yun2026redesign}.
However, a model-level question remains: can generation and decomposition
operate directly in a layer-native visual space, with language controlling what should be
separated, at what granularity, and with which semantic role? In such a space, layers rather
than raw pixels become the atomic units of generation, understanding and editing.

We propose \tsystem{}, a framework that represents images as ordered sets of semantic RGBA
layers. Each layer is a complete, independently addressable visual object that can be
generated, decomposed, extracted, edited, moved, deleted and recomposed. The framework
comprises two models. The Text-to-RGBA (\ttrgba{}) model generates reusable
RGBA assets from text. The Image-to-Layer (\ttil{}) model takes a finished image, a
global task instruction and per-layer prompts, and jointly produces an ordered stack of
complete semantic RGBA layers on a shared canvas. Together, they connect layer generation
and layer decomposition within one representation, allowing visual content to move between
language, finished images and editable structure.
\autoref{fig:teaser} summarizes these two capabilities.

Unlike a fixed segmentation taxonomy, \ttil{} exposes decomposition as an
instruction-addressable operation. Top-level decomposition separates a complete image into
specified semantic layers; recursive decomposition refines a selected layer; and targeted
extraction separates the requested content from the unselected remainder. An external agent can therefore
decide what should become an independent layer, adjust the granularity of an existing
decomposition, or extract a target before editing and recomposition. \tsystem{} provides
reusable, language-addressable object states as an interface for external planning systems.

\begin{figure}[t]
\centering
\includegraphics[width=\linewidth]{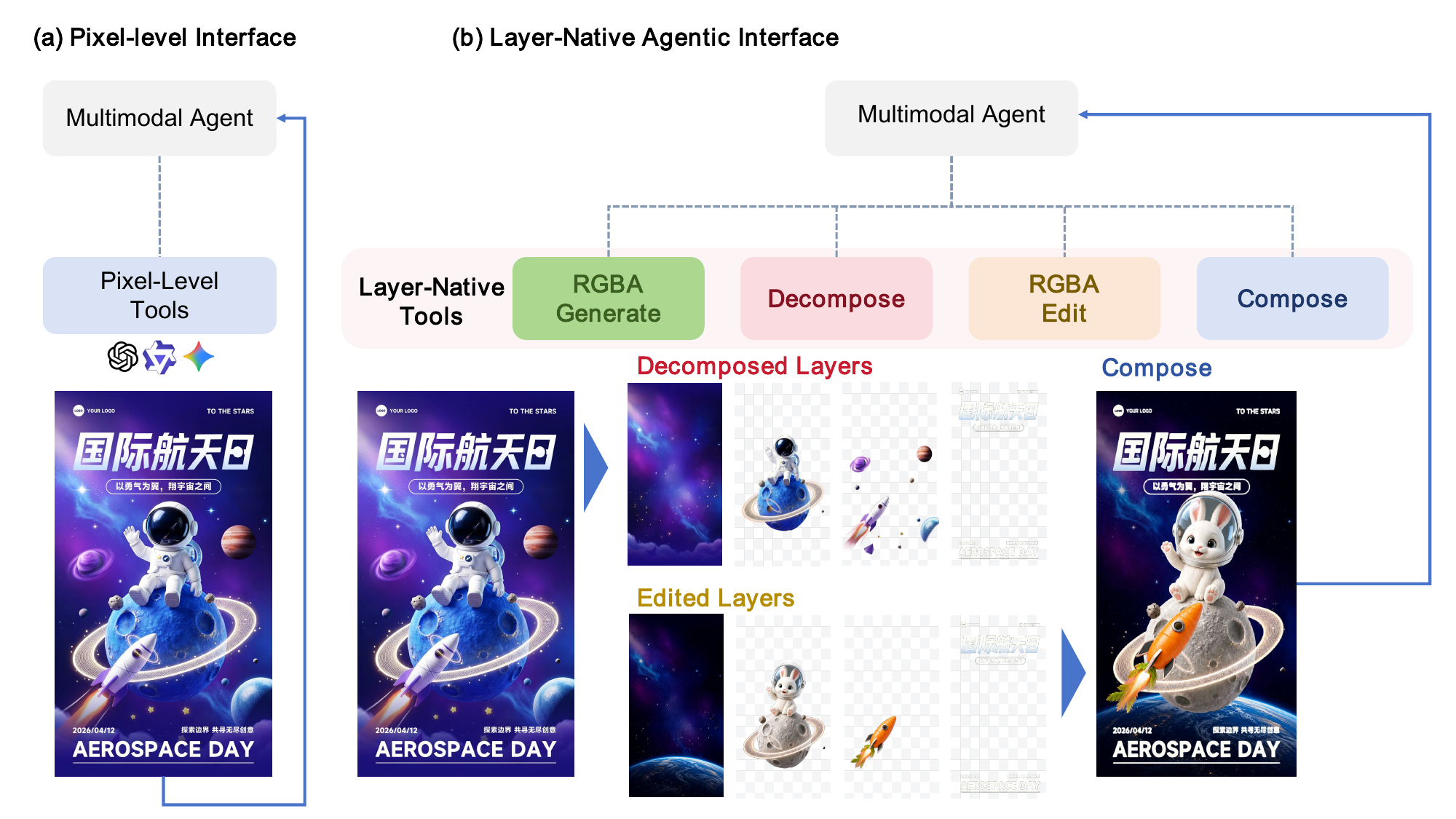}
\caption{\textbf{From pixel generation to agentic layer-native design.}
A pixel-only interface exposes a monolithic raster, requiring object structure to be inferred
again. \tsystem{} instead exposes a persistent ordered state of semantic RGBA layers. An
external design agent can invoke \ttrgba{} generation and \ttil{} decomposition, recursively
refine a returned layer, edit selected layers with other tools, and recompose the state.}
\label{fig:agentic-layer-interface}
\end{figure}

Three design principles underpin this formulation. First, both models treat RGBA as a native
compositional representation rather than merely an output format or post-processing result.
Encoding each object with its own color and alpha
makes it independently placeable and directly composable. Second, \ttil{} preserves the finished image as a
shared visual condition while binding each ordered output to its corresponding semantic
prompt, supporting joint consistency without losing independent control. Third, it learns
complete semantic objects rather than visible-pixel partitions. Visible partitions may recompose
the input accurately yet reveal holes or truncated regions when an occluder is moved;
complete layers retain the hidden content needed for subsequent editing. This establishes a
closed loop in which \ttrgba{} generates semantic assets, \ttil{} converts finished designs
back into language-addressable layers, and selected layers can be refined, modified and
recomposed. \autoref{fig:agentic-layer-interface} illustrates how these operations can
serve an external design agent.

\Needspace*{12\baselineskip}
\paragraph{Contributions.}
This paper makes three contributions:

\begin{itemize}[leftmargin=1.4em,itemsep=1pt,topsep=2pt]
\item A layer-native formulation that treats semantic RGBA layers as the atomic unit of
generation, decomposition and editing. By exposing persistent, independently manipulable
object states, it provides a structured interface over which external design agents can plan
and execute generation, decomposition, refinement, editing and recomposition.
\item Two models that realize this formulation: \ttrgba{} generates reusable
RGBA assets, while \ttil{} provides an instruction-addressable interface for decomposing
finished images into ordered, complete semantic layers at controllable granularity.
\item A systematic evaluation covering per-layer fidelity, transparency, editability and
RGBA asset generation.
\end{itemize}

Compared with Qwen-Image-Layered, \ttil{} reduces per-layer RGB L1 by $37\%$, improves Alpha
Soft IoU by $34\%$, reduces blank-layer generation by $63\%$, and achieves a VLM score of
$20.43$ versus $17.60$. \ttrgba{} reaches the highest CLIP Score
among the models compared ($33.03$). Alpha boundaries and dense text remain challenging.

%% file: sec/2_related_work.tex
\section{Related Work}
\label{sec:related}

\paragraph{RGBA generation.}
Transparency-aware generation has followed two broad strategies. One preserves a pretrained
RGB representation and attaches alpha through latent offsets, auxiliary branches, attention
cues or separate decoders
\citep{zhang2024layerdiffuse,pu2025art}.
The other learns a native RGBA representation for reconstruction, layered generation or
multi-task generation
\citep{wang2025alphavae,liu2025omnipsd,yu2025omnialpha}. Our framework follows the latter
direction but assigns RGBA two roles across two models: \ttrgba{} generates standalone
RGBA assets, whereas \ttil{} represents multiple ordered semantic layers.

\paragraph{Layer decomposition.}
Qwen-Image-Layered jointly generates semantically disentangled RGBA layers and supports
recursive re-decomposition \citep{yin2025qwenimagelayered}; OmniPSD targets text-to-PSD generation and
image-to-PSD recovery \citep{liu2025omnipsd}; and LayerD sequentially extracts elements from
raster graphic designs \citep{suzuki2025layerd}. Related work also provides layered training
data and editability-oriented evaluation protocols
\citep{chen2025prismlayers,tudosiu2024mulan,rowles2026stablelayers}. Qwen-Image-Layered does
not expose per-layer semantic prompts or targeted extraction controlled by text instructions. In
contrast, \ttil{} supports instruction-controlled semantic decomposition by binding a global
task instruction and per-layer prompts to ordered outputs, unifying top-level
decomposition, recursive decomposition and targeted extraction in one interface. It
targets complete semantic layers rather than visible-pixel partitions.

\paragraph{Agentic image editing.}
GenArtist uses a multimodal planner to decompose requests, select visual tools, and verify and
revise intermediate results \citep{wang2024genartist}. More recent systems introduce
layer-aware execution. Agent Banana isolates a mask-localized high-resolution crop, edits it,
and fuses it back, so its ``layer'' is a local execution patch
\citep{ye2026agentbanana}. ReDesign instead grows an editable JSON hierarchy by selecting and
verifying heterogeneous extraction tools, including a layered decomposition model
\citep{yun2026redesign}. These works demonstrate the value of structure in agentic editing,
but focus on planner--executor systems, local patch preservation or multi-tool hierarchy
recovery. In contrast, our native RGBA decomposition model exposes complete semantic layers
as persistent, language-addressable object states, providing external design agents with a
structured interface for planning and executing edits directly in a layer-native visual space.

%% file: sec/3_method.tex
\section{Method}
\label{sec:method}

\subsection{Preliminaries}
\label{sec:method:prelim}

\paragraph{Flow matching.}
Let $z_0$ be a clean sample from the data distribution with a corresponding condition $y$
(text for \ttrgba{}; text and a composite latent for \ttil{}), and let
$z_1 \sim \mathcal{N}(0,I)$ be Gaussian noise. Rectified flow
\citep{liu2023rectifiedflow,lipman2023flowmatching} interpolates between them as
\begin{equation}
z_t = (1-t)\,z_0 + t\,z_1 ,
\qquad t \in [0,1] ,
\label{eq:interp}
\end{equation}
where $t{=}0$ is clean data and $t{=}1$ is noise. The model learns the velocity
$z_1-z_0$ by mean-squared regression \citep{esser2024sd3}, and inference integrates the
learned field from noise to data. All $z$ variables below denote autoencoder latents. We use $i$ for a
layer index, $N$ for the number of target layers, $z^{(i)}$ for layer $i$, and $c$ for the
composite-image condition.

\paragraph{DiffusionNFT.}
DiffusionNFT \citep{zheng2025diffusionnft,li2025uniworld} performs reward-driven
post-training on outputs sampled under the same condition $y$. Let $r_b\in[0,1]$ be
candidate $b$'s group-normalized optimality score, $v_b$ its corresponding target velocity
and $v_{\mathrm{old}}$ the velocity predicted by the
EMA-updated data-collection policy. With the DiffusionNFT implicit-policy parameterization
and $\beta{=}1$, the core objective is
\begin{equation}
\begin{aligned}
v_\theta^{+} &= v_\theta, \qquad
v_\theta^{-} = 2v_{\mathrm{old}}-v_\theta, \\
\mathcal{L}_{\mathrm{NFT}} &= \mathbb{E}_b\!\left[
r_b\lVert v_\theta^{+}-v_b\rVert_2^2
+(1-r_b)\lVert v_\theta^{-}-v_b\rVert_2^2\right].
\end{aligned}
\label{eq:nft-summary}
\end{equation}
High $r_b$ pulls the model toward the sampled target through $v_\theta^{+}$; low $r_b$
pushes it away through the reflected $v_\theta^{-}$. Both branches are parameterized by the
same flow field, so the sampler is unchanged. Reward normalization, adaptive loss weighting
and EMA policy updates follow DiffusionNFT \citep{zheng2025diffusionnft}.

\subsection{RGBA Autoencoder with RGB Latent Alignment}
\label{sec:method:vae}

\paragraph{Channel extension.}
Following recent RGBA autoencoders
\citep{wang2025alphavae,yin2025qwenimagelayered,liu2025omnipsd}, we extend only the
encoder's first and decoder's last convolution from three to four channels. The pretrained
RGB weights are copied, both new alpha filters are
zero-initialized, and the decoder's alpha bias is set to one. The extended model therefore
starts as the original RGB autoencoder with an opaque alpha channel.

\paragraph{Training.}
We train on a mixture of transparent RGBA images and opaque RGB images (with alpha set to
one). The objective combines RGB and alpha reconstruction, perceptual, KL, adversarial and
RGB-latent alignment terms:
\begin{equation}
\begin{aligned}
\mathcal{L}_{\text{VAE}} =\;&
\mathcal{L}^{\text{RGB}}_{1}
+ \lambda_\alpha \mathcal{L}^{\alpha}_{1}
+ \lambda_{\text{P}} \mathcal{L}^{\text{white}}_{\text{LPIPS}}
+ \lambda_{\text{KL}} \mathcal{L}_{\text{KL}} \\
&+ \mathbf{1}_{m\ge m_{\text{GAN}}}\lambda_{\text{GAN}}\rho
\mathcal{L}_{\text{adv}}
+ \lambda_{\text{tea}}
\mathbb{E}_{x\in\mathcal{X}_{\text{RGB}}}
\left\|\operatorname{mode}q_S(z\mid[x,\mathbf{1}])
-\operatorname{mode}q_T(z\mid x)\right\|_2^2 .
\end{aligned}
\label{eq:vae}
\end{equation}
Here $m$ is the optimization step, $m_{\text{GAN}}$ is the adversarial-loss start step and
$\rho$ is the adaptive gradient-norm ratio. The RGBA encoder $q_S$ receives an opaque alpha
channel for an RGB sample $x$, while the frozen teacher $q_T$ receives its three RGB channels. Applying this
alignment only to samples that originate as RGB encourages their encodings to remain close
to the pretrained latent space without directly applying the teacher term to transparent examples.
Autoencoder training uses straight alpha to retain color beneath transparency, whereas
diffusion training represents RGB after compositing over white.

\subsection{Text-to-RGBA Generation}
\label{sec:method:t2rgba}

Training data is an internally curated dataset of text paired with transparent RGBA images,
including design assets, cutout subjects and vector-derived illustrations. Transparent RGBA
images constitute the majority of the training mixture, while opaque images are also included.
Fine-tuning starts at $512$-equivalent area and then continues at $1024$.

Both \ttrgba{} and \ttil{} use the two-stage training pipeline detailed in
\autoref{sec:method:distill}: progressive distillation followed by post-training with
DiffusionNFT \citep{zheng2025diffusionnft}. Their task-specific rewards are specified in
\autoref{eq:task-rewards}. The layering model of \autoref{sec:method:i2l} starts from the
post-trained \ttrgba{} model rather than from the base model. The \ttrgba{} results in
\autoref{sec:exp:t2rgba} are those of the model obtained after distillation and DiffusionNFT
post-training, sampled at $512$-equivalent area.

\subsection{Instruction-Controlled Image-to-Layer Generation}
\label{sec:method:i2l}

\begin{figure}[t]
\centering
\includegraphics[width=\linewidth]{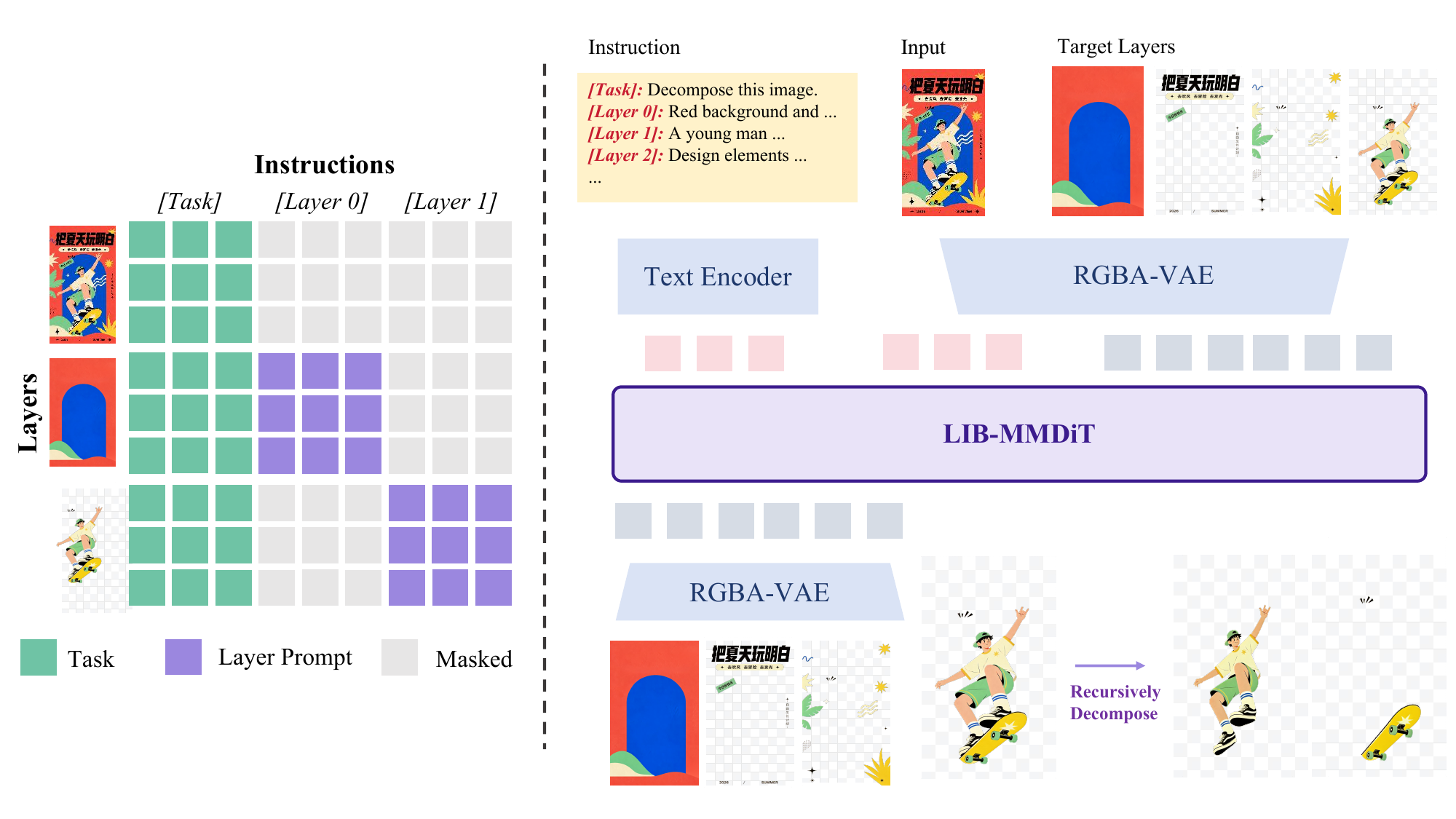}
\caption{\textbf{LIB-MMDiT overview.}
\textbf{Left:} the attention mask broadcasts the global instruction to all image queries
while restricting each layer prompt to its corresponding layer. \textbf{Right:} LIB-MMDiT
jointly generates aligned RGBA layers from an input image and structured layer instructions.}
\label{fig:arch}
\end{figure}

\ttil{} maps a composite image and an instruction to $N$ ordered, complete semantic RGBA
layers on the same canvas. Each target represents a requested semantic role or, for targeted
extraction, the unselected remainder; occluded content is retained. The model is trained so
that alpha-compositing the layers from back to front approximates the input.
Instruction-controlled decomposition creates two
correspondence problems. First,
each semantic description must control the intended output layer rather than the whole
stack. Second, tokens from different layers must remain aligned to the same canvas while
retaining distinct layer identities and order. We call the resulting architecture
\emph{Layer--Instruction Binding MMDiT} (LIB-MMDiT). It combines layer--instruction binding
attention with layer-indexed rotary positions to address the two correspondence problems,
respectively (\autoref{fig:arch}).

\paragraph{Layer--instruction binding attention.}
We assign tag $\tau=-1$ to the global instruction and composite condition, and tag $\tau=i$
to the prompt and image tokens of target layer $i$. An image query reads global text and its
matching prompt, but not prompts assigned to other layers. Image-image attention remains
unrestricted, allowing the condition and all targets to jointly resolve occlusion and
stacking. Text-text attention is also unrestricted; the binding mask applies specifically
when an image query attends to a text key. The global task instruction is therefore
broadcast to the stack, while each layer description is bound only to its intended output.

\paragraph{Layer-indexed 3D rotary positions.}
The base transformer assigns every image token three rotary coordinates $(0,h,w)$, with a
constant first coordinate and spatial row and column. Following the layer-indexed positioning
of Qwen-Image-Layered \citep{yin2025qwenimagelayered}, we reuse that constant coordinate for
layer identity rather than adding another positional axis. Let $p$ denote the rotary
coordinates of an image token, and let $M^{\mathrm{text}}_{qk}$ denote the visibility of a
text key $k$ to an image query $q$, with $\mathcal{I}$ and $\mathcal{T}$ the image- and
text-token sets. The composite uses layer index $-1$, and target layer $i$ uses index $i$,
with $i=0$ the bottom-most layer. Equal $(h,w)$ values align the same canvas location across
the stack, while the first coordinate distinguishes layer order. Text keeps the base model's
position scheme; its layer binding is handled by $M^{\mathrm{text}}$. This reuse introduces
no new RoPE parameters and preserves the base position encoding for a single target at layer
zero.

\Needspace*{5\baselineskip}
Together, the two mechanisms are summarized as
\begin{equation}
\begin{aligned}
p(c_{h,w}) &= (-1,h,w),
& p(z^{(i)}_{h,w}) &= (i,h,w), \quad i=0,\dots,N{-}1, \\
M^{\mathrm{text}}_{qk}
&= \mathbf{1}\!\left[(\tau_k=-1)\ \lor\ (\tau_k=\tau_q)\right],
& q&\in\mathcal{I},\ k\in\mathcal{T}.
\end{aligned}
\label{eq:layer-conditioning}
\end{equation}

\paragraph{Conditioning and training.}
The global instruction, per-layer prompts, encoded composite and target-layer latents enter
one LIB-MMDiT sequence. The composite is kept clean at $t{=}0$, re-injected at every denoising
step and excluded from the loss; only the target layers are noised and supervised, using one
shared timestep per sample. The layer-binding and layer-indexing mechanisms are retained
throughout fine-tuning, distillation and post-training. The exact image-token and text-prompt
packing order is specified in \appref{app:packing}.

\subsection{Building Semantic Layer Trees from PSD Documents}
\label{sec:method:data}

\begin{figure}[t]
\centering
\includegraphics[width=\linewidth]{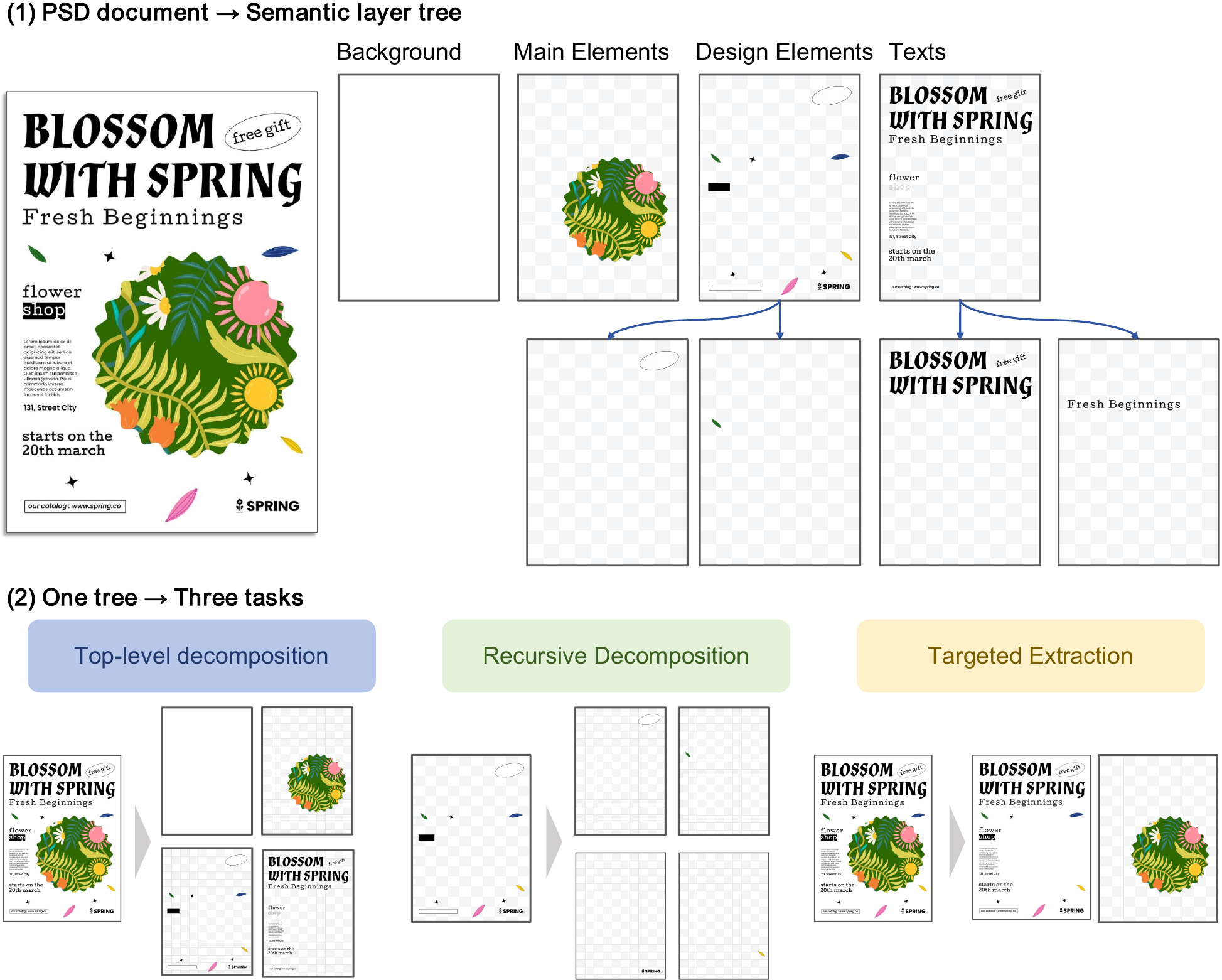}
\caption{\textbf{From one PSD document to three training tasks.} \textbf{(1)} PSD layers
are grouped into a semantic layer tree. \textbf{(2)} Complete semantic layers retain content
behind occluders, unlike visible-pixel partitions. \textbf{(3)} The tree yields top-level
decomposition, recursive decomposition and targeted extraction instances.}
\label{fig:datatree}
\end{figure}

Training \ttil{} requires layer stacks whose layers remain usable when moved or removed.
Synthesizing such stacks from generated assets is circular: their layer statistics inherit
the generator's biases, and any occluded content must itself be generated. We instead use
designer-authored PSD documents, where layer content hidden in the final composite remains
available in the source file. We render the PSD layers, merge related authoring units into
semantic objects and organize them into a hierarchy. A vision-language model assists only
with grouping; all target pixels come from the original PSD layers. \autoref{fig:datatree}
summarizes the construction and its three derived tasks.

\paragraph{Preserving occluded content.}
When constructing a target layer, we preserve its source pixels even where they are hidden
by layers above it. Moving or removing an occluding layer therefore reveals the underlying
content rather than a transparent hole. Visible-only cutouts are not used as training targets.

\paragraph{Tree-derived supervision.}
Each tree node defines a composite and its semantic children. The root provides
top-level decomposition examples, internal nodes provide recursive decomposition examples,
and selected nodes together with the unselected remainder provide targeted extraction examples. The tree
recomposition is used as the image condition so that it is consistent with the retained
targets. The three task types are mixed during fine-tuning.

\FloatBarrier
\subsection{Progressive Distillation and Post-Training}
\label{sec:method:distill}

Emitting several full-resolution layers in one sequence multiplies its length, so sampling
cost dominates inference. After supervised fine-tuning (\autoref{sec:method:i2l}), we use
progressive distillation to reduce sampling cost and post-training with DiffusionNFT to
optimize layer-level output quality. The post-training adapter is merged to obtain the final
layering model.

\paragraph{Empirical choice of trajectory supervision.}
We also experimented with distribution-matching distillation
\citep{yin2024dmd,yin2024dmd2}. In our runs on layered RGBA generation, we observed alpha
predictions tending toward full opacity, misalignment between RGB content and alpha masks,
and weaker condition following
and cross-layer consistency. We therefore adopt progressive adversarial distillation.

\paragraph{Progressive adversarial distillation.}
Following progressive distillation \citep{salimans2022progressive} and the progressive
adversarial formulation of SDXL-Lightning \citep{lin2024sdxllightning}, we train the student to predict a
direction that reaches the flow location obtained after multiple teacher steps. Specifically,
we sample an interval $[t_{k+1},t_k]$ from the coarser student schedule, where
$t_{k+1}<t_k$ under the convention of \autoref{eq:interp}, and add noise to real target
latents to obtain the starting state $\mathbf{Z}_{t_k}$. The frozen teacher then takes $r$ Euler
steps, while the student moves directly to the same time in one step:
\begin{equation}
\begin{aligned}
\widetilde{\mathbf{Z}}^{T}_{t_{k+1}}
  &= \Phi_T^{(r)}(\mathbf{Z}_{t_k};c), \\
\widetilde{\mathbf{Z}}^{S}_{t_{k+1}}
  &= \mathbf{Z}_{t_k}+(t_{k+1}-t_k)
     v_\theta(\mathbf{Z}_{t_k},t_k;c), \\
\mathcal{L}_{\mathrm{distill}}
  &= \frac{1}{N}\sum_{i=0}^{N-1}
     \left\|\widetilde{\mathbf{Z}}^{S,i}_{t_{k+1}}
     -\widetilde{\mathbf{Z}}^{T,i}_{t_{k+1}}\right\|_1
     +\lambda_{\mathrm{adv}}\mathcal{L}_{\mathrm{adv}} .
\end{aligned}
\label{eq:progressive-distill}
\end{equation}
Here $\Phi_T^{(r)}$ denotes the teacher's multi-step update. The clean condition layer and
text form $c$; the condition layer is never noised and is excluded from the sum over $N$
target layers. The endpoint $L_1$ term anchors the student prediction to the teacher's next
flow location. In alternating adversarial updates, a discriminator initialized
from the transformer backbone treats the teacher endpoint as real and the student endpoint
as fake at the same target timestep $t_{k+1}$ and under the same conditioning. This term discourages the
over-smoothed endpoints produced by distance regression alone. Distillation yields an
eight-step student, which remains the evaluation schedule after post-training.

\paragraph{Task-specific rewards.}
For DiffusionNFT post-training, the two models use different reward pipelines. For
\ttrgba{}, every decoded RGBA sample is
evaluated by three reward heads. Let $a(\hat{x},y)$ denote the alpha-following score, which
reads the generated alpha channel and the prompt's transparent/opaque label, rewarding
transparency when a cut-out is requested and full coverage when an opaque image is
requested. Let $s_Q(\hat{x},y)$ denote the CLIP-style prompt--image cosine similarity
computed using embeddings from Qwen3-VL \citep{bai2025qwen3vl}.
Following the MLLM implicit feedback of UniWorld-v2 \citep{li2025uniworld}, let
$s_M(\hat{x},y)$ be the logit-based score
from Qwen3.5-9B \citep{qwen2026qwen35}. We prompt the frozen MLLM to rate the white-composited
asset against its text condition on a scale from $0$ to $5$, apply a softmax to the output
logits of the six score tokens, and use their normalized expected value,
$s_M=\frac{1}{5}\sum_{k=0}^{5}k\,\operatorname{softmax}(\ell)_k$.

For \ttil{}, every decoded layer $\hat{x}^{(i)}$ is paired by its ordered slot with the
ground-truth RGBA layer $x^{(i)}$ on the same canvas. The first reward head is the mean
absolute error over all four RGBA channels and all pixels. The second is a frozen AlexNet
LPIPS network \citep{zhang2018lpips} applied only to the three RGB channels, without an
alpha mask:
\[
d_1^{(i)}=\operatorname{mean}\!\left|\hat{x}^{(i)}-x^{(i)}\right|,
\qquad
d_P^{(i)}=\operatorname{LPIPS}_{\mathrm{Alex}}
\!\left(\hat{x}^{(i)}_{\mathrm{RGB}},x^{(i)}_{\mathrm{RGB}}\right).
\]
The rewards are
\begin{equation}
\begin{aligned}
R_{\mathrm{T2RGBA}}(\hat{x},y)
  &= a(\hat{x},y)\left[\lambda_Q s_Q(\hat{x},y)+\lambda_M s_M(\hat{x},y)\right], \\
R_{\mathrm{I2L}}(\widehat{\mathbf{x}},\mathbf{x})
  &= -\frac{1}{N}\sum_{i=0}^{N-1}
     \frac{\lambda_1 d_1^{(i)}+\lambda_P d_P^{(i)}}{\lambda_1+\lambda_P} .
\end{aligned}
\label{eq:task-rewards}
\end{equation}
The outer alpha-following term is a gate: a sample cannot compensate for incorrect opacity
merely by receiving high embedding or MLLM scores. For \ttil{}, the two explicit
reconstruction distances are first combined within each layer and then averaged equally
across target layers; the
stable configuration uses equal $L_1$ and LPIPS weights. We do not multiply rewards across
layers and use no reward on the recomposed image. Alpha errors therefore affect the RGBA
$L_1$ head but never enter the perceptual network.

%% file: sec/4_experiment.tex
\Needspace*{9\baselineskip}
\section{Evaluation}
\label{sec:exp}

\subsection{Setup}
\label{sec:exp:protocol}

We evaluate I2L by per-layer fidelity and editability, and T2RGBA by text alignment,
reference resemblance, and alpha quality.

\paragraph{Image-to-Layer.}
We use $512$ fixed-seed Crello designs \citep{yamaguchi2021canvasvae}, pairing each rendered
design with its source stack; no Crello data are used for training. Our final checkpoint runs
at $512$-equivalent image area, eight steps, and no classifier-free guidance. We request four
layers to match Qwen-Image-Layered,
which runs through its official script \citep{yin2025qwenimagelayered} at the recommended
$640$ pixels with its optional global-caption input.

LayerD \citep{suzuki2025layerd} supplies stack alignment and per-layer RGB/alpha metrics;
Stable-Layers \citep{rowles2026stablelayers} supplies reference-free editability and the
structural rubric. Definitions and resampling rules are in \appref{app:eval};
each $\pm$ is one standard deviation across samples.

\paragraph{Text-to-RGBA.}
From an internally curated held-out set, we sample $512$ captioned layers at seed $42$ and
generate one $512{\times}512$ asset per caption. The evaluation captions, their matched
source layers and the real-layer FID reference set do not overlap the training data. CLIP
Score uses all outputs; FID \citep{heusel2017fid} compares their white composites with
$1000$ real layers. Alpha MSE,
SAD, and white-composite LPIPS use the $479$ locatable source layers at their native
resolutions; availability is fixed before inference and shared by all methods. We compare
LayerDiffuse \citep{zhang2024layerdiffuse} and OmniAlpha
\citep{yu2025omnialpha}.

\subsection{Layer Decomposition Quality}
\label{sec:exp:i2l}

\begin{table}[H]
\centering
\caption{\textbf{Per-layer quality on 512 Crello designs.} LayerD dynamic-time-warping
alignment is applied before scoring; RGB L1 uses binary reference-alpha support.}
\label{tab:i2l}
\small
\setlength{\heavyrulewidth}{1.1pt}
\begin{tabular}{lcc}
\toprule
Method & RGB L1 $\downarrow$ & Alpha Soft IoU $\uparrow$ \\
\midrule
Qwen-Image-Layered ($4$ layers) & $0.2014${\scriptsize$\pm0.0937$} & $0.5454${\scriptsize$\pm0.1577$} \\
\rowcolor{RowBlue}
\tshort{}-\ttil{} ($4$ layers) & $\mathbf{0.1264}${\scriptsize$\pm0.0980$} & $\mathbf{0.7325}${\scriptsize$\pm0.1622$} \\
\bottomrule
\end{tabular}
\end{table}

At matched layer count, \autoref{tab:i2l} shows a $37\%$ relative reduction in per-layer
RGB L1 ($0.2014 \rightarrow 0.1264$) and a $34\%$ relative increase in Alpha Soft IoU
($0.5454 \rightarrow 0.7325$). Thus, after stack alignment, our layers more closely match
both the RGB content and alpha support of their references.

\begin{table}[H]
\centering
\caption{\textbf{Editability on 512 Crello designs.} Stable-Layers reference-free metrics;
Bad Layers counts blank and glazed layers per sample, and the percentage is per emitted slot.}
\label{tab:i2l-edit}
\small
\setlength{\heavyrulewidth}{1.1pt}
\begin{tabular}{lcccc}
\toprule
Method & Bad Layers $\downarrow$ & Blank $\downarrow$ & Glaze $\downarrow$ & Feat.\ Dist.\ $\uparrow$ \\
\midrule
Qwen-Image-Layered ($4$ layers) & $1.69$ ($42.3\%$) & $0.35$ & $1.34$ & $0.658$ \\
\rowcolor{RowBlue}
\tshort{}-\ttil{} ($4$ layers) & $\mathbf{1.32}$ ($\mathbf{33.0\%}$) & $\mathbf{0.13}$ & $\mathbf{1.19}$ & $\mathbf{0.690}$ \\
\bottomrule
\end{tabular}
\end{table}

The reference-free metrics in \autoref{tab:i2l-edit} are consistent with the per-layer fidelity
results in \autoref{tab:i2l}. Blank outputs fall by $63\%$ ($0.35 \rightarrow 0.13$), and
Feature Distribution rises from $0.658$ to $0.690$, suggesting less redundant content.
Glazed outputs also fall from $1.34$ to $1.19$, reducing Bad Layers from $1.69$ to $1.32$.

\begin{table}[H]
\centering
\caption{\textbf{VLM-based evaluation of four-layer decomposition} following the
Stable-Layers protocol. Five dimensions are scored from $0$ to $5$; values are mean
$\pm$ standard deviation across $512$ samples.
The total is out of $25$, with its normalized value in parentheses.}
\label{tab:i2l-vlm}
\small
\setlength{\heavyrulewidth}{1.1pt}
\setlength{\tabcolsep}{2.8pt}
\begin{tabular}{lcccccc}
\toprule
Method & \makecell{Semantic\\separation} & \makecell{Alpha\\cleanliness} &
\makecell{Background\\inpainting} & \makecell{Content\\distribution} &
\makecell{Content\\validity} & \makecell{Total\\(normalized)} \\
\midrule
Qwen-Image-Layered & $3.00{\pm}0.99$ & $\mathbf{3.33}{\pm}1.59$ & $3.88{\pm}1.64$ &
$3.60{\pm}1.17$ & $3.78{\pm}1.39$ & $17.60{\pm}4.97$ ($0.704$) \\
\rowcolor{RowBlue}
\tshort{}-\ttil{} & $\mathbf{3.83}{\pm}0.86$ & $2.90{\pm}1.11$ &
$\mathbf{4.44}{\pm}1.08$ & $\mathbf{4.55}{\pm}0.67$ & $\mathbf{4.71}{\pm}0.70$ &
$\mathbf{20.43}{\pm}2.91$ ($\mathbf{0.817}$) \\
\bottomrule
\end{tabular}
\end{table}

The VLM evaluation in \autoref{tab:i2l-vlm} shows improvements in four dimensions and a
decline in alpha cleanliness. Content distribution ($+0.95$), content validity ($+0.93$),
and semantic separation ($+0.83$)
account for most of the $2.83$-point total improvement; background inpainting adds $0.56$.
The latter is compatible with the effect expected from supervision using complete objects
in real documents, although we do not isolate that effect here. Alpha
cleanliness is the only lower dimension ($2.90$ versus $3.33$) and is discussed in
\autoref{sec:discussion}.

\begin{figure}[!t]
\centering
\includegraphics[width=\linewidth]{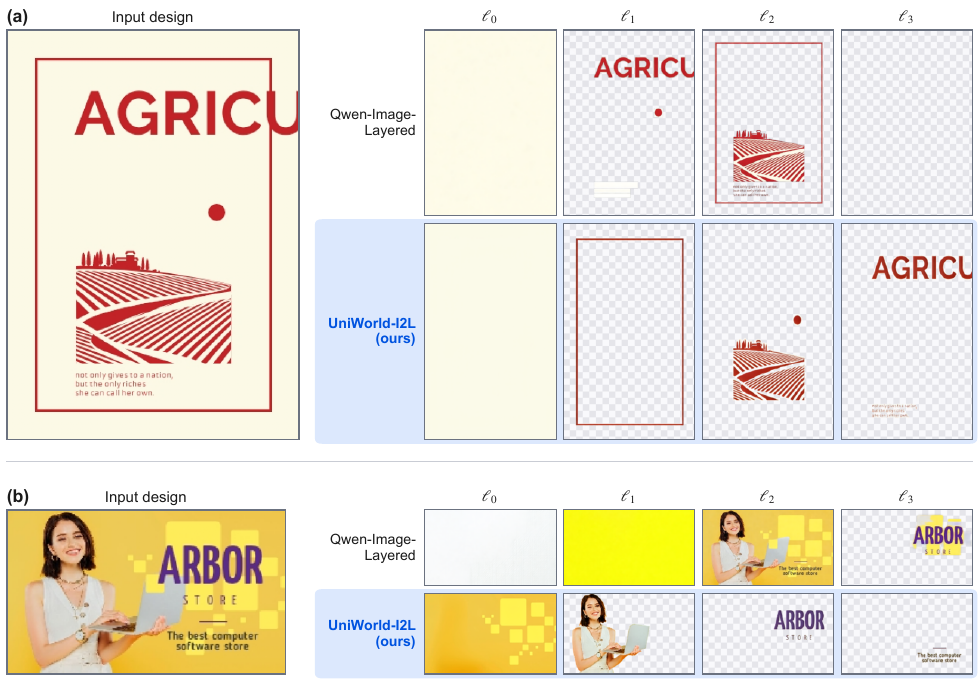}
\caption{\textbf{Qualitative comparison of per-layer decomposition with Qwen-Image-Layered
on two Crello designs.}
For each design, the Qwen-Image-Layered row appears above ours; each row shows four predicted
RGBA layers, $\ell_0$ through $\ell_3$, ordered back to front. Checkerboard denotes
transparency.}
\label{fig:qual-i2l}
\end{figure}

\paragraph{Qualitative comparison.}
\autoref{fig:qual-i2l} shows the same structural pattern on individual designs, while
\autoref{fig:recursive} demonstrates that a returned layer can be decomposed again by a
second call.

\begin{figure*}[p]
\centering
\includegraphics[page=1,width=\textwidth]{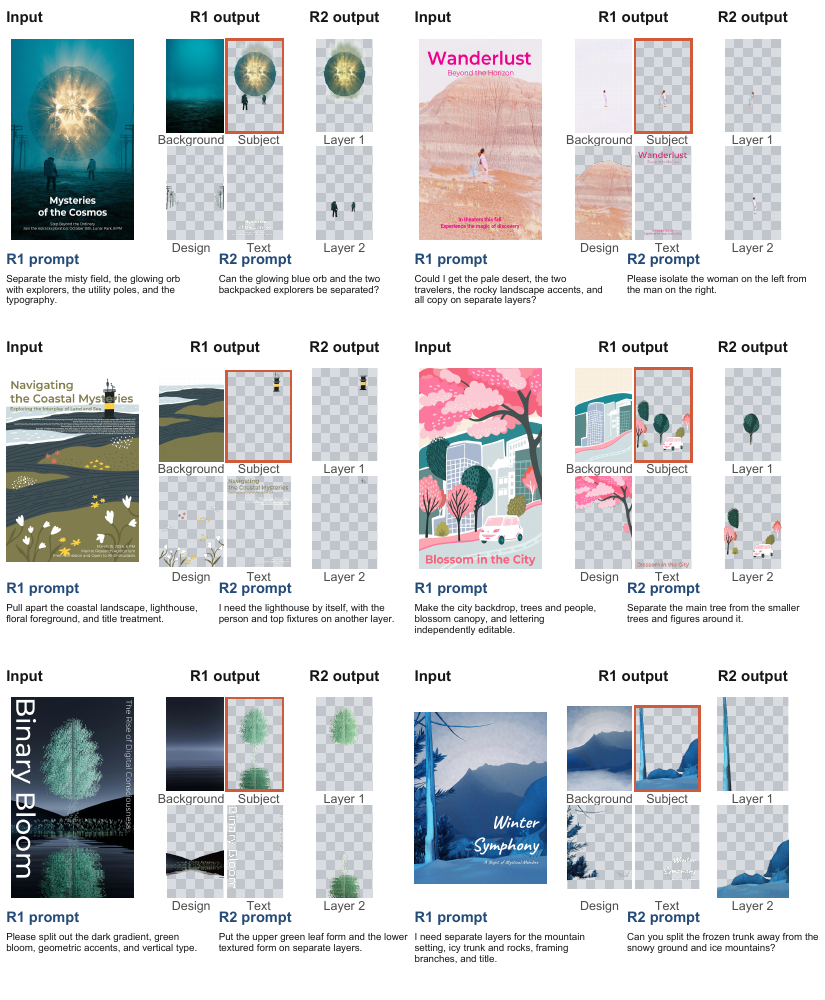}
\caption{\textbf{Qualitative analysis of instruction-controlled I2L decomposition.}
For each input poster, the first round (R1) follows the user prompt shown below the example
to decompose the image into four semantic, position-aligned RGBA layers: background,
subject, design, and text. The outlined subject layer becomes the input image for the second
round (R2). R2 then follows its own user-specified prompt to decompose
that selected layer into two finer-grained layers, again preserving position and alpha.}
\label{fig:recursive}
\end{figure*}

\begin{figure*}[p]
\ContinuedFloat
\centering
\includegraphics[page=2,width=\textwidth]{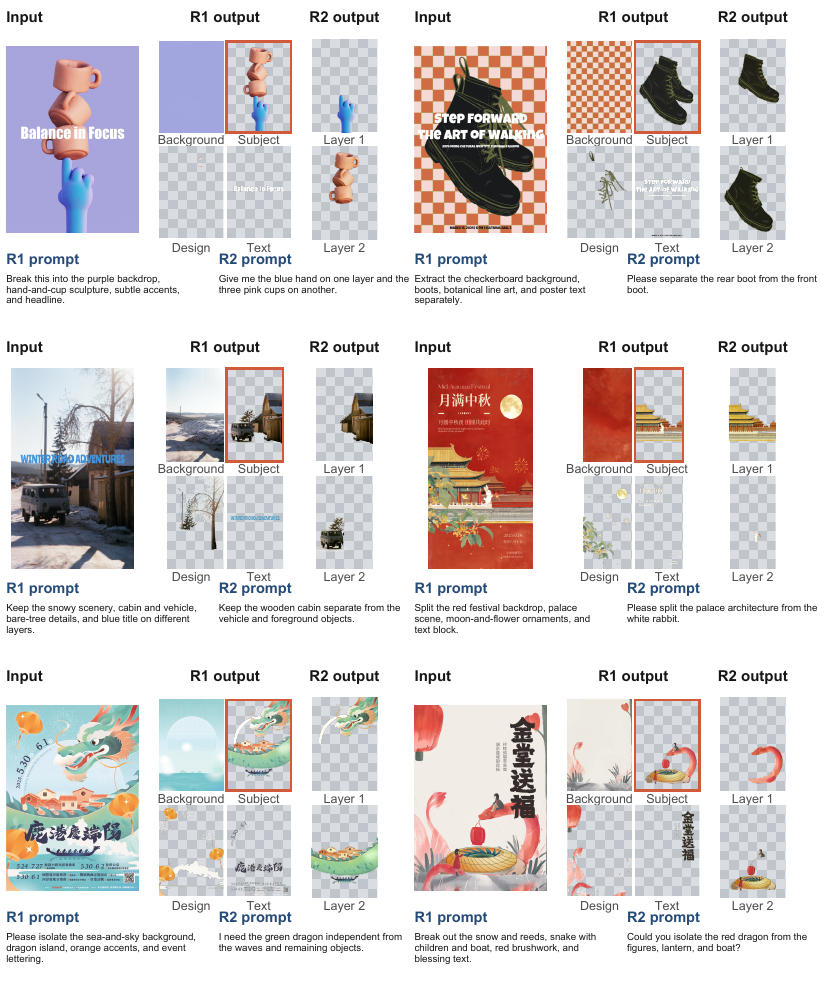}
\caption[]{\textbf{Qualitative analysis of instruction-controlled I2L decomposition (continued).}}
\end{figure*}

\begin{figure*}[p]
\ContinuedFloat
\centering
\includegraphics[page=3,width=\textwidth]{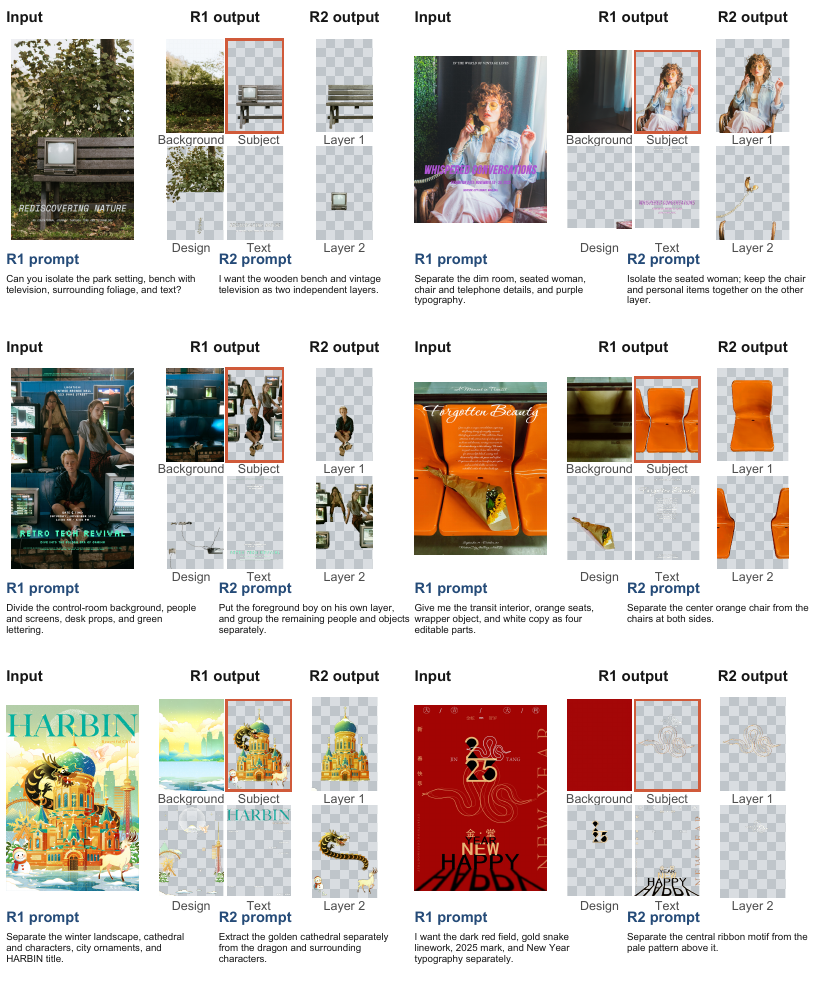}
\vspace{-18pt}
\caption[]{\textbf{Qualitative analysis of instruction-controlled I2L decomposition (continued).}}
\end{figure*}

\subsection{Text-to-RGBA Generation}
\label{sec:exp:t2rgba}

\begin{table}[H]
\centering
\caption{\textbf{Text-to-RGBA generation on 512 design-element prompts.} FID and CLIP Score
use all $512$ generations, with FID computed against $1000$ real layers. Alpha MSE, SAD
\citep{levin2008matting}, and LPIPS use the $479$ prompts with matched reference layers,
evaluated at the reference resolution (\autoref{sec:exp:protocol}, \appref{app:eval}).
OmniAlpha uses its released reinforcement-learning LoRA at $1024{\times}1024$, $50$ steps,
and true guidance $4.0$.}
\label{tab:t2rgba}
\small
\setlength{\heavyrulewidth}{1.1pt}
\begin{tabular}{lccccc}
\toprule
Method & FID $\downarrow$ & CLIP Score $\uparrow$ & Alpha MSE $\downarrow$ & SAD $\downarrow$ & LPIPS (white) $\downarrow$ \\
\midrule
LayerDiffuse (SDXL) & $214.47$ & $29.22${\scriptsize$\pm4.51$} & $0.495${\scriptsize$\pm0.322$} & $1{,}281{,}170$ & $0.724${\scriptsize$\pm0.205$} \\
OmniAlpha & $\mathbf{87.86}$ & $31.00${\scriptsize$\pm7.02$} & $\mathbf{0.406}${\scriptsize$\pm0.305$} & $\mathbf{1{,}049{,}631}$ & $\mathbf{0.462}${\scriptsize$\pm0.244$} \\
\rowcolor{RowBlue}
\tshort{}-\ttrgba{} & $117.14$ & $\mathbf{33.03}${\scriptsize$\pm5.29$} & $0.413${\scriptsize$\pm0.382$} & $1{,}083{,}441$ & $0.537${\scriptsize$\pm0.245$} \\
\bottomrule
\end{tabular}
\end{table}

\begin{figure}[H]
\centering
\includegraphics[width=\linewidth]{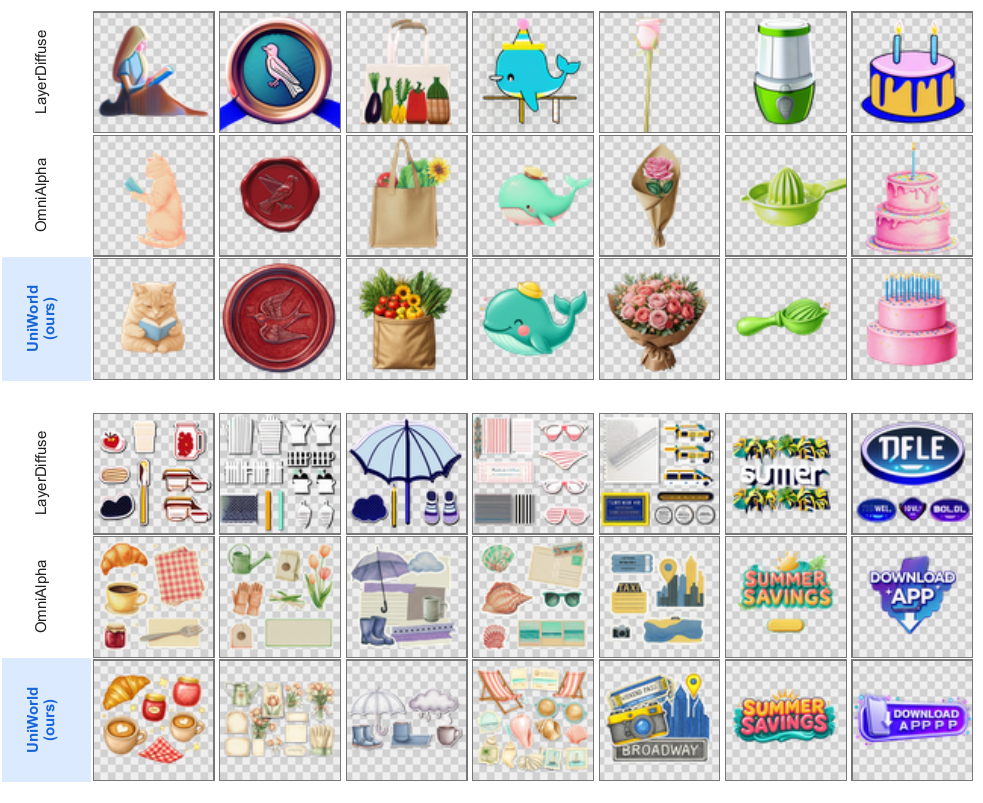}
\caption{\textbf{Text-to-RGBA against LayerDiffuse and OmniAlpha.} The top panel contains
seven single-resource prompts and the bottom panel seven multi-resource prompts. Each column
uses the same prompt for all three models; the full English prompts are listed in
\appref{app:prompts}.}
\label{fig:qual-t2rgba}
\end{figure}

\Needspace{4\baselineskip}
In \autoref{tab:t2rgba}, the improvement over LayerDiffuse is consistent: FID falls by $45\%$,
CLIP Score rises by $13\%$, and the three paired measures improve by $15$--$26\%$. Our model
has the highest mean CLIP Score ($33.03$
versus $31.00$, $+6.5\%$), whereas OmniAlpha leads on FID ($87.86$ versus $117.14$) and
white-composite LPIPS ($0.462$ versus $0.537$). Its Alpha MSE and SAD are also lower, but by
small margins ($0.406$ versus $0.413$ and $1{,}049{,}631$ versus $1{,}083{,}441$,
respectively).

\Needspace{8\baselineskip}
These metrics answer different questions. Prompt--image evaluation ranges from embedding
alignment to knowledge-informed semantic criteria \citep{radford2021clip,niu2025wise}; in
our protocol, CLIP Score serves as the alignment proxy, while FID and LPIPS compare outputs
with reference assets at the distribution and instance levels, respectively. Within this
protocol, OmniAlpha
better matches the reference appearance distribution and achieves slightly lower Alpha MSE
and SAD, while our model follows the text more closely on average.
\autoref{fig:qual-t2rgba} complements these aggregate
metrics with side-by-side generations for selected single-resource and multi-resource design
briefs.

\FloatBarrier

%% file: sec/discussion.tex
\section{Limitations}
\label{sec:discussion}

\paragraph{Alpha edges.}
\autoref{tab:i2l-vlm} reports lower alpha cleanliness for \tshort{}-\ttil{} than for
Qwen-Image-Layered ($2.90$ versus $3.33$). Improving fine alpha boundaries remains future
work.

\paragraph{Dense typography, and Chinese text in particular.}
Complex glyphs and longer passages produce missing strokes, wrong characters and unstable
layouts. Typography-intensive design will require stronger text generation in the underlying
model.

%% file: sec/5_conclusion.tex
\section{Conclusion}
\label{sec:conclusion}

\tsystem{} treats semantic RGBA layers as native units of generation and interaction rather
than outputs of a post-processing pipeline. T2RGBA generates reusable standalone RGBA
assets, while I2L decomposes finished images into ordered, complete semantic layers through
top-level decomposition,
recursive refinement and targeted extraction. These layers form persistent,
language-addressable object states that external design agents can select, manipulate with
other tools and recompose.

Our evaluations show that I2L improves per-layer fidelity and usability over
Qwen-Image-Layered, while T2RGBA achieves the highest mean CLIP Score among the compared
models. Alpha-boundary quality and dense typography remain the primary limitations observed
in our evaluations.

Next, we will scale data, model capacity and compute to pretrain a unified model for native
RGBA generation and understanding, extending its scope from RGBA asset generation and image
decomposition to layer insertion and replacement, layout-constrained composition and
iterative multi-turn editing.

%% file: sec/contributions.tex
\section{Contributions}
\label{sec:contributions}

\textbf{Contributors}:
Zongjian Li, Chenxu Bai, Chen Chen, Haoxiang Sun, Shaodong Wang, Feize Wu,
Shenghai Yuan, Bin Lin, Zheyuan Liu, Yuwei Niu.

\textbf{Project Leads}: Zhiyuan Yan, Li Yuan$^{\dagger}$
\blfootnote{$^{\dagger}$Corresponding author.}

%% file: sec/6_appendix.tex
\section{Evaluation Details}
\label{app:eval}

\paragraph{Per-layer metrics (LayerD).}
Predicted and reference layer stacks are treated as ordered sequences and aligned by the
LayerD dynamic-time-warping edit protocol, which may merge adjacent layers before scoring to
accommodate granularity differences. Under that alignment, each matched pair is compared at
the smaller of the two resolutions (downsampling only). For $N$ matched pairs with predicted
alpha $A^{pred}$, reference alpha $A^{gt}$, binary reference support
$S_i(h,w)=\mathbf{1}[A^{gt}_i(h,w)>0]$, predicted RGB $R$ and reference RGB $G$, all with
RGB and alpha values in $[0,1]$,
\begin{align}
\mathcal{L}_{\text{RGB}} &= \frac{1}{N}\sum_{i=0}^{N-1}
\frac{\sum_{h,w}S_i(h,w)\,\frac{1}{3}
\lVert R_i(h,w)-G_i(h,w)\rVert_1}
     {\sum_{h,w}S_i(h,w)} , \\[2pt]
\text{Soft IoU}_\alpha &= \frac{1}{N}\sum_{i=0}^{N-1}
\frac{\sum_{h,w}\min\!\big(A^{pred}_i(h,w), A^{gt}_i(h,w)\big)}
     {\sum_{h,w}\max\!\big(A^{pred}_i(h,w), A^{gt}_i(h,w)\big)} ,
\end{align}
both with values in $[0,1]$. RGB L1 is weighted by the binary reference support, so a layer is judged
where the reference alpha is nonzero, and Alpha Soft IoU uses continuous alpha rather than a
binarized mask, which makes it sensitive to semi-transparent regions.

\paragraph{Editability metrics.}
Following Stable-Layers, we compute these metrics using only the predicted layers. A layer is
\emph{blank} when
its mean alpha is below $0.01$. A layer is \emph{glazed} when its mean alpha lies in
$(0.1, 0.9)$ and its RGB standard deviation exceeds $0.05$ --- a diffuse semi-transparent
layer containing residual color, which limits its independent editability. Bad Layers is the
sum of the two
counts. Feature Distribution extracts a DINOv2-base feature per layer (the spatial mean of
the last hidden state) and reports $1 - \frac{2}{n(n-1)}\sum_{i<j}\cos(f_i,f_j)$, so a higher
value means the layers of one output are less alike.

\paragraph{VLM judging.}
GPT-5.6-terra receives the input design, the alpha composite of the predicted
stack and a labeled contact sheet of the predicted layers, and returns five integers in
$[0,5]$: semantic separation (does each foreground layer contain a single complete element
rather than a fragment), alpha cleanliness (are transparent
boundaries free of color residue, halos and haze), background inpainting (is the region
behind the foreground plausibly restored in $\ell_0$), content distribution (is content
spread meaningfully across layers rather than concentrated in one), and content validity
(does the recomposition match the input, with no missing content, blank layers or noise
layers).

\paragraph{\ttrgba{} metrics.}
FID uses Inception-v3 features at $299{\times}299$ after compositing on white and compares
$512$ generated samples with a reference distribution constructed from $1000$ real layers.
CLIP Score is
the cosine similarity between the generated asset and its prompt under CLIP ViT-B/32, over
the same $512$ generations. Alpha
MSE, SAD and white-composite LPIPS are computed at the native resolution of the corresponding
reference layer, using the $479$ prompts with successfully matched source layers
(\autoref{sec:exp:protocol}); all three methods are scored on that same subset. Alpha
MSE is the mean squared error between the predicted and reference alpha channels
on $[0,1]$, and SAD \citep{levin2008matting} is the classical matting statistic
$\sum_{i,j}\lvert A^{pred}_{i,j} - A^{gt}_{i,j}\rvert$ over the reference layer's own pixels.
White-composite LPIPS uses an AlexNet backbone.

\section{Sequence Construction}
\label{app:packing}

\paragraph{Token packing.}
Tokens are packed in layer-major order --- every token of layer $0$, then every token of layer
$1$, and so on --- rather than interleaving layers at each spatial position. Combined with a
shared crop and resize across the layers of a sample, this makes a given $(h,w)$ refer to the
same canvas location in every layer. Text is packed as a global \texttt{instruction}
followed, for each target layer $i$, by \texttt{Layer $i$:\ <short> --- <detailed>}; the
explicit index is what binds a description to a layer slot and lets the layer count be set at
inference.

\section{Prompts for the Qualitative \ttrgba{} Comparison}
\label{app:prompts}

\autoref{fig:qual-t2rgba} uses seven single-resource prompts in its top panel and seven
multi-resource prompts in its bottom panel. The prompts below follow the figure's left-to-right
column order.

\input{sec/selected_rgba_prompts_appendix}

\enlargethispage{2\baselineskip}

%% file: sec/selected_rgba_prompts_appendix.tex

\paragraph{Single-resource prompts (top panel).}

\noindent\textbf{Prompt.} one peach colored cat reading a tiny sky blue book only, peach, sky blue, vanilla palette, refined editorial gouache texture, delicate colored-pencil detail, soft controlled highlights, crisp clean silhouette, subject occupies 70 percent of canvas, centered, transparent background\par

\noindent\textbf{Prompt.} one complete round red wax seal stamp with embossed postal bird emblem, full seal visible only, postal red, indigo, aged paper palette, refined editorial gouache texture, delicate colored-pencil detail, soft controlled highlights, crisp clean silhouette, subject occupies 70 percent of canvas, centered, transparent background\par

\noindent\textbf{Prompt.} one kraft canvas market tote filled with colorful vegetables only, tomato red, leafy green, sunflower yellow palette, refined editorial gouache texture, delicate colored-pencil detail, soft controlled highlights, crisp clean silhouette, subject occupies 70 percent of canvas, centered, transparent background\par

\noindent\textbf{Prompt.} one kawaii turquoise whale wearing a tiny yellow sailor hat only, turquoise, lemon, bubblegum pink palette, refined editorial gouache texture, delicate colored-pencil detail, soft controlled highlights, crisp clean silhouette, subject occupies 70 percent of canvas, centered, transparent background\par

\noindent\textbf{Prompt.} one rose pink flower bouquet wrapped in kraft paper only, rose pink, leaf green, kraft brown palette, refined editorial gouache texture, delicate colored-pencil detail, soft controlled highlights, crisp clean silhouette, subject occupies 70 percent of canvas, centered, transparent background\par

\noindent\textbf{Prompt.} one complete lime green hand citrus juicer with ribbed bowl and short handle, full juicer visible only, lime green, orange, clean white palette, refined editorial gouache texture, delicate colored-pencil detail, soft controlled highlights, crisp clean silhouette, subject occupies 70 percent of canvas, centered, transparent background\par

\noindent\textbf{Prompt.} one two tier birthday cake with pink frosting and blue candles only, confetti pink, bright blue, golden yellow palette, refined editorial gouache texture, delicate colored-pencil detail, soft controlled highlights, crisp clean silhouette, subject occupies 70 percent of canvas, centered, transparent background\par

\paragraph{Multi-resource prompts (bottom panel).}

\noindent\textbf{Prompt.} Create a polished cozy breakfast sticker sheet as a cohesive hand-journal and scrapbook design asset. Include croissant, jam jar, coffee cup, gingham napkin, tiny sparkles. Use a butter yellow, strawberry red, oat beige palette, charming hand-drawn gouache and colored-pencil texture, delicate paper grain, clean white sticker borders, subtle layered paper shadows, balanced spacing, and varied scale. Arrange the pieces as one compact editorial sticker cluster, isolated on a truly transparent background, no mockup, no watermark, no extra text, crisp cutout edges, premium commercial design quality.\par

\noindent\textbf{Prompt.} Create a polished spring gardening scrapbook kit as a cohesive hand-journal and scrapbook design asset. Include watering can, seed packet, tulips, garden gloves, paper labels. Use a sage green, peach, cream palette, charming hand-drawn gouache and colored-pencil texture, delicate paper grain, clean white sticker borders, subtle layered paper shadows, balanced spacing, and varied scale. Arrange the pieces as one compact editorial sticker cluster, isolated on a truly transparent background, no mockup, no watermark, no extra text, crisp cutout edges, premium commercial design quality.\par

\noindent\textbf{Prompt.} Create a polished rainy day journaling cluster as a cohesive hand-journal and scrapbook design asset. Include clear umbrella, rain boots, cloud, tea mug, dotted washi tape. Use a dusty blue, lavender, warm gray palette, charming hand-drawn gouache and colored-pencil texture, delicate paper grain, clean white sticker borders, subtle layered paper shadows, balanced spacing, and varied scale. Arrange the pieces as one compact editorial sticker cluster, isolated on a truly transparent background, no mockup, no watermark, no extra text, crisp cutout edges, premium commercial design quality.\par

\noindent\textbf{Prompt.} Create a polished seaside vacation ephemera set as a cohesive hand-journal and scrapbook design asset. Include shells, striped deck chair, postcard, sunglasses, postage stamps. Use a aqua, coral, sand palette, charming hand-drawn gouache and colored-pencil texture, delicate paper grain, clean white sticker borders, subtle layered paper shadows, balanced spacing, and varied scale. Arrange the pieces as one compact editorial sticker cluster, isolated on a truly transparent background, no mockup, no watermark, no extra text, crisp cutout edges, premium commercial design quality.\par

\noindent\textbf{Prompt.} Create a polished city weekend travel ephemera as a cohesive hand-journal and scrapbook design asset. Include subway ticket, camera, skyline, map pin, street sign. Use a cobalt, taxi yellow, charcoal palette, charming hand-drawn gouache and colored-pencil texture, delicate paper grain, clean white sticker borders, subtle layered paper shadows, balanced spacing, and varied scale. Arrange the pieces as one compact editorial sticker cluster, isolated on a truly transparent background, no mockup, no watermark, no extra text, crisp cutout edges, premium commercial design quality.\par

\noindent\textbf{Prompt.} Design a high-converting summer promotion decorative type for a modern e-commerce campaign. The exact main lettering must read "SUMMER SAVINGS". Use a aqua, coral, sunny yellow palette with sun, wave, tropical leaves, bubbly dimensional letters; create bold custom display typography, clear hierarchy, dimensional highlights, clean shadows, energetic but controlled ornament, and excellent readability at thumbnail size. Isolate one complete button or badge asset on a truly transparent background, no product photo, no mockup, no watermark, no extra words, crisp alpha edges, premium agency-level commercial finish.\par

\noindent\textbf{Prompt.} Design a high-converting app download promotional badge for a modern e-commerce campaign. The exact main lettering must read "DOWNLOAD APP". Use a gradient blue, purple, white palette with smartphone, downward arrow, pixel sparkles; create bold custom display typography, clear hierarchy, dimensional highlights, clean shadows, energetic but controlled ornament, and excellent readability at thumbnail size. Isolate one complete button or badge asset on a truly transparent background, no product photo, no mockup, no watermark, no extra words, crisp alpha edges, premium agency-level commercial finish.\par

%% file: main.bib
@article{lin2025uniworld,
  title   = {UniWorld: High-Resolution Semantic Encoders for Unified Visual Understanding and Generation},
  author  = {Lin, Bin and Li, Zongjian and Cheng, Xinhua and Niu, Yuwei and Ye, Yang and He, Xianyi and Yuan, Shenghai and Yu, Wangbo and Wang, Shaodong and Ge, Yunyang and others},
  journal = {arXiv preprint arXiv:2506.03147},
  year    = {2025}
}

@article{diao2026sensenova,
  title   = {SenseNova-U1: Unifying Multimodal Understanding and Generation with {NEO-unify} Architecture},
  author  = {Diao, Haiwen and Wu, Penghao and Deng, Hanming and Wang, Jiahao and Bai, Shihao and Wu, Silei and Fan, Weichen and Ye, Wenjie and Tong, Wenwen and Fan, Xiangyu and others},
  journal = {arXiv preprint arXiv:2605.12500},
  year    = {2026}
}

@inproceedings{Yan_2026_CVPR,
  author    = {Yan, Zhiyuan and Lin, Kaiqing and Li, Zongjian and Ye, Junyan and Han, Hui and Wang, Haochen and Wang, Zhendong and Lin, Bin and Li, Hao and Xiao, Xinyan and Wang, Jingdong and Wang, Haifeng and Yuan, Li},
  title     = {Unified Multimodal Models as Auto-Encoders},
  booktitle = {Proceedings of the IEEE/CVF Conference on Computer Vision and Pattern Recognition (CVPR)},
  month     = {June},
  year      = {2026},
  pages     = {41903--41912}
}

@article{bai2025qwen3vl,
  title   = {Qwen3-VL Technical Report},
  author  = {Bai, Shuai and Cai, Yuxuan and Chen, Ruizhe and Chen, Keqin and Chen, Xionghui and Cheng, Zesen and Deng, Lianghao and Ding, Wei and Gao, Chang and Ge, Chunjiang and Ge, Wenbin and Guo, Zhifang and Huang, Qidong and Huang, Jie and Huang, Fei and others},
  journal = {arXiv preprint arXiv:2511.21631},
  year    = {2025}
}

@misc{qwen2026qwen35,
  title        = {{Qwen3.5-9B}},
  author       = {{Qwen Team}},
  year         = {2026},
  howpublished = {Hugging Face model card},
  url          = {https://huggingface.co/Qwen/Qwen3.5-9B}
}

@article{zhang2024layerdiffuse,
  title   = {Transparent Image Layer Diffusion using Latent Transparency},
  author  = {Zhang, Lvmin and Agrawala, Maneesh},
  journal = {arXiv preprint arXiv:2402.17113},
  year    = {2024}
}

@article{wang2025alphavae,
  title   = {AlphaVAE: Unified End-to-End RGBA Image Reconstruction and Generation with Alpha-Aware Representation Learning},
  author  = {Wang, Zile and Yu, Hao and Zhan, Jiabo and Yuan, Chun},
  journal = {arXiv preprint arXiv:2507.09308},
  year    = {2025}
}

@article{yu2025omnialpha,
  title   = {OmniAlpha: Aligning Transparency-Aware Generation via Multi-Task Unified Reinforcement Learning},
  author  = {Yu, Hao and Wang, Jinglin and Zhan, Jiabo and Chen, Rui and Wang, Zile and Zhang, Huaisong and Li, Hongyu and Chen, Xinrui and Wei, Yongxian and Yuan, Chun},
  journal = {arXiv preprint arXiv:2511.20211},
  year    = {2025}
}

@inproceedings{pu2025art,
  title     = {ART: Anonymous Region Transformer for Variable Multi-Layer Transparent Image Generation},
  author    = {Pu, Yifan and Zhao, Yiming and Tang, Zhicong and Yin, Ruihong and Ye, Haoxing and Yuan, Yuhui and Chen, Dong and Bao, Jianmin and Zhang, Sirui and Wang, Yanbin and Liang, Lin and Wang, Lijuan and Li, Ji and Li, Xiu and Lian, Zhouhui and others},
  booktitle = {IEEE/CVF Conference on Computer Vision and Pattern Recognition (CVPR)},
  year      = {2025}
}

@article{yin2025qwenimagelayered,
  title   = {Qwen-Image-Layered: Towards Inherent Editability via Layer Decomposition},
  author  = {Yin, Shengming and Zhang, Zekai and Tang, Zecheng and Gao, Kaiyuan and Xu, Xiao and Yan, Kun and Li, Jiahao and Chen, Yilei and Chen, Yuxiang and Shum, Heung-Yeung and Ni, Lionel M. and Zhou, Jingren and Lin, Junyang and Wu, Chenfei},
  journal = {arXiv preprint arXiv:2512.15603},
  year    = {2025}
}

@inproceedings{suzuki2025layerd,
  title         = {LayerD: Decomposing Raster Graphic Designs into Layers},
  author        = {Suzuki, Tomoyuki and Liu, Kang-Jun and Inoue, Naoto and Yamaguchi, Kota},
  booktitle     = {IEEE/CVF International Conference on Computer Vision (ICCV)},
  year          = {2025},
  eprint        = {2509.25134},
  archivePrefix = {arXiv}
}

@article{rowles2026stablelayers,
  title   = {Stable-Layers: Fine-Tuning Image Layer Decomposition Models with VLM-Scored Reinforcement Learning},
  author  = {Rowles, Ciara and Adithyan, Reshinth and Pinnaparaju, Nikhil and Voleti, Vikram and Boss, Mark},
  journal = {arXiv preprint arXiv:2605.30257},
  year    = {2026}
}

@article{liu2025omnipsd,
  title   = {OmniPSD: Layered PSD Generation with Diffusion Transformer},
  author  = {Liu, Cheng and Song, Yiren and Wang, Haofan and Shou, Mike Zheng},
  journal = {arXiv preprint arXiv:2512.09247},
  year    = {2025}
}

@article{chen2025prismlayers,
  title   = {PrismLayers: Open Data for High-Quality Multi-Layer Transparent Image Generative Models},
  author  = {Chen, Junwen and Jiang, Heyang and Wang, Yanbin and Wu, Keming and Li, Ji and Zhang, Chao and Yanai, Keiji and Chen, Dong and Yuan, Yuhui},
  journal = {arXiv preprint arXiv:2505.22523},
  year    = {2025}
}

@article{tudosiu2024mulan,
  title   = {MULAN: A Multi Layer Annotated Dataset for Controllable Text-to-Image Generation},
  author  = {Tudosiu, Petru-Daniel and Yang, Yongxin and Zhang, Shifeng and Chen, Fei and McDonagh, Steven and Lampouras, Gerasimos and Iacobacci, Ignacio and Parisot, Sarah},
  journal = {arXiv preprint arXiv:2404.02790},
  year    = {2024}
}

@inproceedings{yamaguchi2021canvasvae,
  title     = {CanvasVAE: Learning to Generate Vector Graphic Documents},
  author    = {Yamaguchi, Kota},
  booktitle = {IEEE/CVF International Conference on Computer Vision (ICCV)},
  year      = {2021}
}

@inproceedings{brooks2023instructpix2pix,
  title     = {InstructPix2Pix: Learning to Follow Image Editing Instructions},
  author    = {Brooks, Tim and Holynski, Aleksander and Efros, Alexei A.},
  booktitle = {IEEE/CVF Conference on Computer Vision and Pattern Recognition (CVPR)},
  pages     = {18392--18402},
  year      = {2023}
}

@inproceedings{couairon2023diffedit,
  title     = {DiffEdit: Diffusion-Based Semantic Image Editing with Mask Guidance},
  author    = {Couairon, Guillaume and Verbeek, Jakob and Schwenk, Holger and Cord, Matthieu},
  booktitle = {International Conference on Learning Representations (ICLR)},
  year      = {2023}
}

@inproceedings{wang2024genartist,
  title     = {GenArtist: Multimodal LLM as an Agent for Unified Image Generation and Editing},
  author    = {Wang, Zhenyu and Li, Aoxue and Li, Zhenguo and Liu, Xihui},
  booktitle = {Advances in Neural Information Processing Systems (NeurIPS)},
  volume    = {37},
  year      = {2024}
}

@article{ye2026agentbanana,
  title   = {Agent Banana: High-Fidelity Image Editing with Agentic Thinking and Tooling},
  author  = {Ye, Ruijie and Zhang, Jiayi and Liu, Zhuoxin and Zhu, Zihao and Yang, Siyuan and Li, Li and others},
  journal = {arXiv preprint arXiv:2602.09084},
  year    = {2026}
}

@inproceedings{yun2026redesign,
  title     = {{ReDesign}: Recovering Editable Design Structures from Images via Agentic Decomposition},
  author    = {Yun, Jooyeol and Park, Jintae and Lim, Hyesu and Hyung, Junha and Chung, Hyungjin and Choo, Jaegul},
  booktitle = {European Conference on Computer Vision (ECCV)},
  year      = {2026}
}

@article{zheng2024birefnet,
  title   = {Bilateral Reference for High-Resolution Dichotomous Image Segmentation},
  author  = {Zheng, Peng and Gao, Dehong and Fan, Deng-Ping and Liu, Li and Laaksonen, Jorma and Ouyang, Wanli and Sebe, Nicu},
  journal = {CAAI Artificial Intelligence Research},
  volume  = {3},
  pages   = {9150038},
  year    = {2024}
}

@inproceedings{qin2022dis,
  title     = {Highly Accurate Dichotomous Image Segmentation},
  author    = {Qin, Xuebin and Dai, Hang and Hu, Xiaobin and Fan, Deng-Ping and Shao, Ling and Van Gool, Luc},
  booktitle = {European Conference on Computer Vision (ECCV)},
  year      = {2022}
}

@inproceedings{liu2023rectifiedflow,
  title     = {Flow Straight and Fast: Learning to Generate and Transfer Data with Rectified Flow},
  author    = {Liu, Xingchao and Gong, Chengyue and Liu, Qiang},
  booktitle = {International Conference on Learning Representations (ICLR)},
  year      = {2023}
}

@inproceedings{lipman2023flowmatching,
  title     = {Flow Matching for Generative Modeling},
  author    = {Lipman, Yaron and Chen, Ricky T. Q. and Ben-Hamu, Heli and Nickel, Maximilian and Le, Matt},
  booktitle = {International Conference on Learning Representations (ICLR)},
  year      = {2023}
}

@inproceedings{esser2024sd3,
  title     = {Scaling Rectified Flow Transformers for High-Resolution Image Synthesis},
  author    = {Esser, Patrick and Kulal, Sumith and Blattmann, Andreas and Entezari, Rahim and M{\"u}ller, Jonas and Saini, Harry and Levi, Yam and Lorenz, Dominik and Sauer, Axel and Boesel, Frederic and Podell, Dustin and Dockhorn, Tim and English, Zion and Lacey, Kyle and Goodwin, Alex and others},
  booktitle = {International Conference on Machine Learning (ICML)},
  year      = {2024}
}

@inproceedings{salimans2022progressive,
  title     = {Progressive Distillation for Fast Sampling of Diffusion Models},
  author    = {Salimans, Tim and Ho, Jonathan},
  booktitle = {International Conference on Learning Representations (ICLR)},
  year      = {2022}
}

@article{lin2024sdxllightning,
  title   = {{SDXL-Lightning}: Progressive Adversarial Diffusion Distillation},
  author  = {Lin, Shanchuan and Wang, Anran and Yang, Xiao},
  journal = {arXiv preprint arXiv:2402.13929},
  year    = {2024}
}

@inproceedings{yin2024dmd,
  title     = {One-step Diffusion with Distribution Matching Distillation},
  author    = {Yin, Tianwei and Gharbi, Micha{\"e}l and Zhang, Richard and Shechtman, Eli and Durand, Fredo and Freeman, William T. and Park, Taesung},
  booktitle = {IEEE/CVF Conference on Computer Vision and Pattern Recognition (CVPR)},
  year      = {2024}
}

@inproceedings{yin2024dmd2,
  title     = {Improved Distribution Matching Distillation for Fast Image Synthesis},
  author    = {Yin, Tianwei and Gharbi, Micha{\"e}l and Park, Taesung and Zhang, Richard and Shechtman, Eli and Durand, Fredo and Freeman, William T.},
  booktitle = {Advances in Neural Information Processing Systems (NeurIPS)},
  year      = {2024}
}

@inproceedings{zheng2025diffusionnft,
  title     = {DiffusionNFT: Online Diffusion Reinforcement with Forward Process},
  author    = {Zheng, Kaiwen and Chen, Huayu and Ye, Haotian and Wang, Haoxiang and Zhang, Qinsheng and Jiang, Kai and Su, Hang and Ermon, Stefano and Zhu, Jun and Liu, Ming-Yu},
  booktitle = {International Conference on Learning Representations (ICLR)},
  year      = {2026},
  note      = {Oral}
}

@article{li2025uniworld,
  title   = {UniWorld-v2: Reinforce Image Editing with Diffusion Negative-Aware Finetuning and MLLM Implicit Feedback},
  author  = {Li, Zongjian and Liu, Zheyuan and Zhang, Qihui and Lin, Bin and Wu, Feize and Yuan, Shenghai and Yan, Zhiyuan and Ye, Yang and Yu, Wangbo and Niu, Yuwei and others},
  journal = {arXiv preprint arXiv:2510.16888},
  year    = {2025}
}

@article{niu2025wise,
  title   = {{WISE}: A World Knowledge-Informed Semantic Evaluation for Text-to-Image Generation},
  author  = {Niu, Yuwei and Ning, Munan and Zheng, Mengren and Jin, Weiyang and Lin, Bin and Jin, Peng and Liao, Jiaqi and Feng, Chaoran and Ning, Kunpeng and Zhu, Bin and others},
  journal = {arXiv preprint arXiv:2503.07265},
  year    = {2025}
}

@inproceedings{radford2021clip,
  title     = {Learning Transferable Visual Models From Natural Language Supervision},
  author    = {Radford, Alec and Kim, Jong Wook and Hallacy, Chris and Ramesh, Aditya and Goh, Gabriel and Agarwal, Sandhini and Sastry, Girish and Askell, Amanda and Mishkin, Pamela and Clark, Jack and Krueger, Gretchen and Sutskever, Ilya},
  booktitle = {International Conference on Machine Learning (ICML)},
  year      = {2021}
}

@inproceedings{heusel2017fid,
  title     = {GANs Trained by a Two Time-Scale Update Rule Converge to a Local Nash Equilibrium},
  author    = {Heusel, Martin and Ramsauer, Hubert and Unterthiner, Thomas and Nessler, Bernhard and Hochreiter, Sepp},
  booktitle = {Advances in Neural Information Processing Systems (NeurIPS)},
  year      = {2017}
}

@inproceedings{zhang2018lpips,
  title     = {The Unreasonable Effectiveness of Deep Features as a Perceptual Metric},
  author    = {Zhang, Richard and Isola, Phillip and Efros, Alexei A. and Shechtman, Eli and Wang, Oliver},
  booktitle = {IEEE/CVF Conference on Computer Vision and Pattern Recognition (CVPR)},
  year      = {2018}
}

@inproceedings{zhu2017semanticamodal,
  title     = {Semantic Amodal Segmentation},
  author    = {Zhu, Yan and Tian, Yuandong and Metaxas, Dimitris and Doll{\'a}r, Piotr},
  booktitle = {IEEE/CVF Conference on Computer Vision and Pattern Recognition (CVPR)},
  year      = {2017}
}

@inproceedings{ozguroglu2024pix2gestalt,
  title     = {pix2gestalt: Amodal Segmentation by Synthesizing Wholes},
  author    = {Ozguroglu, Ege and Liu, Ruoshi and Sur{\'i}s, D{\'i}dac and Chen, Dian and Dave, Achal and Tokmakov, Pavel and Vondrick, Carl},
  booktitle = {IEEE/CVF Conference on Computer Vision and Pattern Recognition (CVPR)},
  year      = {2024}
}

@article{levin2008matting,
  title   = {A Closed-Form Solution to Natural Image Matting},
  author  = {Levin, Anat and Lischinski, Dani and Weiss, Yair},
  journal = {IEEE Transactions on Pattern Analysis and Machine Intelligence},
  volume  = {30},
  number  = {2},
  pages   = {228--242},
  year    = {2008}
}
